\documentclass[conference]{IEEEtran}
\IEEEoverridecommandlockouts  % needed when the author/affiliation block runs wide

\usepackage[T1]{fontenc}
\usepackage[utf8]{inputenc}
\usepackage{cite}             % IEEE-preferred numeric citation handling
\usepackage{amsmath,amssymb,amsfonts}
\usepackage{algorithmic}
\usepackage{graphicx}
\usepackage{booktabs}
\usepackage{textcomp}
\usepackage{xcolor}
\usepackage{url}
\usepackage[hidelinks]{hyperref}  % load last

\newcommand{\histX}{X_{u}}                 % user u's purchase/transaction history
\newcommand{\pattern}{\rho}                % a transaction pattern (set of >=1 items)
\newcommand{\patternset}{\mathcal{P}}      % deduplicated set of unique patterns
\newcommand{\minsup}{\sigma_{\min}}        % minimum support for pattern mining

\newcommand{\schema}{\mathcal{A}}                       % closed-set attribute schema
\newcommand{\schemadem}{\mathcal{A}^{\mathrm{dem}}}     % demographic (7)
\newcommand{\schemapb}{\mathcal{A}^{\mathrm{pb}}}       % psychographic+behavioral (4)
\newcommand{\schemale}{\mathcal{A}^{\mathrm{le}}}       % life-event (8)
\newcommand{\profile}{Y_{u}}                            % hybrid profile of user u
\newcommand{\Yclosed}{Y^{\mathrm{closed}}_{u}}          % Layer 1: closed-set
\newcommand{\Yopen}{Y^{\mathrm{open}}_{u}}              % Layer 2: open-set free-text
\newcommand{\freeattr}{d}                               % a free-text attribute
\newcommand{\prior}[1]{\pi(#1)}
\newcommand{\judge}{J}                                  % LLM-as-a-judge predictor (symmetric)
\begin{document}

% 第一案 (working). Subtitle-tuning option in outline.md "Open structural decisions".
\title{From ``Who Is This User?'' to\\
``What Does This Purchase Mean?'':\\
A Deployed Pipeline for\\
Semantic User Profiling at Bank Scale}

% single-blind: authors named at submission.
\author{\IEEEauthorblockN{Ryota Mitsuhashi\textsuperscript{\dag}, Tetsuro Morimura\textsuperscript{\dag}, Hirotake Ito}
\IEEEauthorblockA{CyberAgent, Tokyo, Japan}
\thanks{\textsuperscript{\dag}These authors contributed equally to this work.}}

\maketitle

\begin{abstract}
Per-user LLM inference on transaction histories binds the inference budget linearly to user count, which becomes prohibitive at applied scale.
We re-cast attribute inference from per-user to per-transaction-pattern.
The pipeline runs in three phases: \emph{Resolve} abstracts item names with optional web grounding, \emph{Profile} infers attributes for each frequent pattern, and \emph{Tag} clusters free-text attributes into a queryable database.
In Profile, a single LLM call per pattern emits predefined categorical labels, free-text attributes, and per-attribute prevalence estimates.
Because inference runs over patterns rather than users, the budget grows with the pattern count rather than the user count.
On the public Open e-commerce corpus, the database is statistically indistinguishable from an LLM that reads each user's raw history directly in AUC across the evaluated attributes, and the prevalence estimates carry discriminative signal between positive and negative users.
The pipeline is deployed at a major Japanese bank profiling on the order of tens of millions of users, with close to a three-order-of-magnitude reduction in LLM inference targets versus a per-user pipeline.
The code is publicly available on \url{https://github.com/CyberAgentAILab/profiling-agent-open-ecommerce}.
\end{abstract}


\begin{figure*}[!t]
\centering
\includegraphics[width=\textwidth]{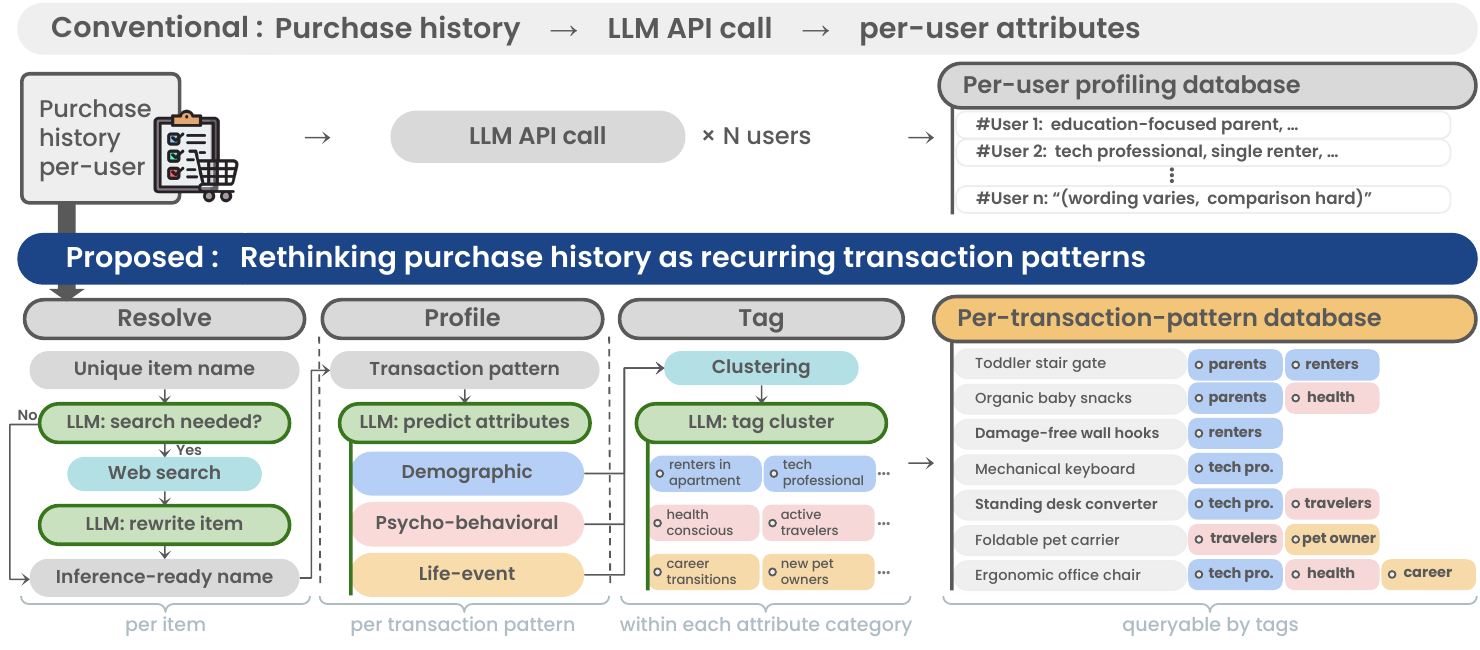}
\caption{Per-user vs.\ per-transaction-pattern profiling. Conventional pipelines (top) call the LLM once per user. The proposed pipeline (bottom) groups three phases (Section~\ref{sec:method}): \emph{Resolve} abstracts item names with optional web grounding; \emph{Profile} jointly emits closed-set labels, free-text attributes, and per-attribute priors across three categories (demographic, psychographic-behavioral, life-event); \emph{Tag} clusters free-text attributes within each attribute category and names each cluster as a tag. The resulting attribute database is queryable by tags. For clarity the figure traces only the free-text path on single-item patterns; closed-set labels appear in the same Profile output, and multi-item patterns mined in Profile share the same formulation.}
\label{fig:teaser}
\end{figure*}

\section{Introduction}
\label{sec:intro}

% ¶1 Motivation — examples-first; long-tail; no closed-set enumerates in advance; existence proof from §7
Marketing campaigns~\cite{yan09_bt_empirical}, financial product personalization, and content recommendation routinely act on natural-language descriptions of a user such as ``an education-focused parent of a school-age child'' or ``recently had a child.''
Useful descriptions span demographic situations, behaviors, and life events, and they form a long tail that no closed-set list of categories enumerates in advance: the IAB Tech Lab Audience Taxonomy~\cite{iab20_audience_taxonomy} fixes roughly a thousand nodes and still leaves out most of the relational, situational descriptions an operator wants to act on.
That such descriptions are operationally useful is established in current practice: at the deploying bank of Section~\ref{sec:deployment}, staff manually read individual users' transactions and write short natural-language descriptions of exactly this form — free text, not a label index or a numerical vector~\cite{ramos24_nl_user_profiles,shi25_personas_ecommerce}.
We formalize these descriptions as the open-set, free-text side of the user profile in Section~\ref{sec:formulation}.

% ¶2 Gap — earlier work did not yield NL descriptions; LLMs make NL generation
%    straightforward; the obvious realization, adopted by essentially all recent work,
%    is per-user prompting [bridge to ¶3]. Pre-LLM detail deferred to §2 ¶1.
Earlier work on inferring user attributes from purchase or transaction histories did not produce descriptions of this form, returning instead either an assignment to a fixed schema or a learned vector embedding (Section~\ref{sec:related}).
Large language models change what is straightforward to generate: given a user's items, the model can be prompted to read them and emit a short natural-language description.
The natural realization, and the one adopted by essentially all recent work that takes this route~\cite{shi25_personas_ecommerce,wan24_tnt_llm,shestov25_llm4es}, is to prompt the LLM once per user.
This realization inherits structural costs that bind at applied scale.

% ¶3 Key idea — three walls of per-user inference; pattern shift removes all three by
%    construction; order-of-magnitude cost footnote (no deployment-specific absolutes);
%    one forward reference to Fig.~\ref{fig:saturation}.
Per-user LLM inference faces three structural walls.
First, the initial build cost scales linearly in the number of users; at the scale of a major bank with tens of millions of customers, a single per-user LLM call at current public-API prices runs into an order of magnitude operators do not budget for as routine\footnote{At a representative public-API rate of a few US dollars per million output tokens and one to a few thousand tokens per user, the per-user cost lies in $10^{-3}$ to $10^{-2}$ US dollars, giving a build cost of $10^{4}$ to $10^{5}$ US dollars per full pass at tens of millions of users; the pass must be repeated when histories change. Rates follow the public {API} price list for a representative cost-efficient hosted model (gpt-5-mini; \url{https://developers.openai.com/api/docs/pricing}, accessed September 2026), and token counts are public-model-card estimates.}.
Second, when the LLM reads each user in isolation, the wording varies across users even when their behavior overlaps, so cross-user comparison, aggregation, and search become unreliable without a separate canonicalization step.
Third, user transaction logs grow continuously, so keeping the database current requires re-inferring every user whose log has changed; the same linear factor that gates the build gates each refresh.

Our central move is to ask not ``who is this user?'' but ``what does this purchase pattern mean?'', running the LLM over the finite, cross-user-shared set of frequent transaction patterns rather than over users.
The inference budget then scales with the count of unique patterns rather than with the user count, which on public data plateaus once the user count exceeds the inverse of the support threshold (Fig.~\ref{fig:saturation}) and at the bank's scale reduces inference targets by close to three orders of magnitude (Section~\ref{sec:deployment}).
Two users matched to the same pattern receive the same attributes, so cross-user comparison and aggregation are well-defined without a canonicalization pass; refresh runs only on patterns that did not exist before, and users matching previously-seen patterns reuse cached attributes.

% ¶4 Approach — joint single-pass attribute inference; two complementary evaluation tracks
%    (selectable asymmetric, tag-based symmetric); deployment + Amazon reproduction.
The pipeline, sketched in Figure~\ref{fig:teaser}, performs a joint LLM call per transaction pattern, emitting closed-set labels across three attribute categories (demographic, psychographic-behavioral, life-event), open-set free-text attributes within each category, and a per-attribute prior in $[0,1]$ that estimates prevalence among purchasers of the pattern.
The per-transaction-pattern outputs are aggregated to per-user profiles, and the open-set side is clustered and tagged within each category to form a queryable attribute database whose query unit is a tag rather than a free-text string.
The pipeline is deployed at a major Japanese bank for advertising and audience analysis (Section~\ref{sec:deployment}) and reproduced on the public Open e-commerce corpus~\cite{berke24_open_ecommerce} with a self-report survey (Section~\ref{sec:setup}--Section~\ref{sec:results}).

% ¶5 Contributions (C1 formulation+method / C2 evaluation / C3 deployment), each
%     mapped to one of ICDM Applied Track's 5 review axes; no release statement.
We make three contributions.
\textbf{C1 (Formulation and Method, originality and technical quality):} we re-cast user-attribute inference from transaction histories as a per-transaction-pattern rather than per-user problem (Section~\ref{sec:formulation}) and instantiate it as a pipeline that emits closed-set labels, open-set free-text attributes, and per-attribute priors across three attribute categories in one joint LLM call per pattern (Section~\ref{sec:method}); both the formulation and its realization are part of the methodological contribution.
\textbf{C2 (Evaluation, technical quality and clarity):} we evaluate information preservation on the public Open e-commerce corpus through two complementary settings (Section~\ref{sec:setup}; Section~\ref{sec:results}), report on the discriminative-signal validity of the per-attribute prior (Section~\ref{sec:results}), and characterize the resulting tag inventory and coverage (Section~\ref{sec:results}).
\textbf{C3 (Deployment, relevance and significance):} we report a deployment at a major Japanese bank, with aggregate system metrics expressed as ratios rather than absolutes and the operational role the pipeline plays (Section~\ref{sec:deployment}).

\section{Related Work}
\label{sec:related}

% ¶1 Pre-LLM line: closed-set structured prediction + SSL/foundation embeddings.
%    Uniform critique: per-user; varying output form (closed labels OR opaque embeddings).
%    LLM-based per-user profiling is deferred to ¶2.
Pre-LLM work on inferring attributes from purchase or transaction histories splits into two lines.
Structured prediction over a closed demographic schema: \cite{wang16_cart_demographics} for retail baskets, \cite{kim19_etna} adding task-specific attention, \cite{hu07_demographic_browsing} on browsing logs, and \cite{resheff17_fusing_transaction} fusing heterogeneous transaction fields.
Learned vector embeddings of the transaction sequence: \cite{babaev22_coles} (contrastive sequence embeddings), \cite{skalski23_foundation_purchasing} (a GPT-style foundation model pretrained on five billion card transactions across 180 banks), and \cite{baldassini18_client2vec}.
Both lines run inference per user and emit either a small fixed schema or an embedding vector, neither of which is a readable, queryable natural-language description of the kind Section~\ref{sec:intro} motivates; LLM-based per-user profiling, which does produce such descriptions, is treated separately below.

% ¶2 LLM-based per-user profiling — natural successor to ¶1. shi25 / wan24 / shestov25
%    each get an explicit differentiation. Our departure is dual (output form + inference unit).
%    Do NOT re-list the 3-wall framing here; cost line is one sentence.
LLM-based per-user profiling is the natural successor: \cite{shi25_personas_ecommerce} generates short natural-language personas and propagates them across users with a graph-based affinity step; \cite{wan24_tnt_llm} mines a label taxonomy with iterative LLM prompting and distils it into a lightweight classifier for scale; and \cite{shestov25_llm4es}, the most direct LLM competitor on transaction data, fine-tunes an LLM by next-token prediction over textualized event sequences and extracts user embeddings from the hidden states.
The three differ in output form (persona label, distilled classifier, embedding vector) but share two assumptions that make cost linear in the user base: a single output representation, and per-user inference.
Natural-language profiles for recommendation~\cite{ramos24_nl_user_profiles,mysore23_lace_editable_profiles,park23_generative_agents} and the broader LLM-for-recommendation line~\cite{geng22_p5,bao23_tallrec} follow the same per-user pattern.
We depart on both dimensions simultaneously, generating a hybrid output (closed-set labels, open-set free-text attributes, and per-attribute priors) within a single LLM call performed at the level of transaction patterns; the cost and consistency consequences are spelled out in Section~\ref{sec:intro} and Section~\ref{sec:formulation}.

% ¶3 Life-event detection: narrow standalone task with engineered features and a small discrete event set.
%    Position: life events are one of three attribute categories in our unified profile,
%    so direct numerical comparison is not the appropriate frame.
A separate body of work addresses life-event detection from bank transactions as a standalone task: \cite{decaigny20_life_events} predicts four discrete events (moving, childbirth, a new relationship, a breakup) from sixty million debit transactions using gradient boosting on engineered RFM and pseudo-social-network features; \cite{beyerdiaz24_improved_life_event} is the follow-up quantifying business impact; and \cite{diclemente18_credit_lifestyles} clusters credit-card sequences into a small set of lifestyle archetypes.
This line treats the task one event at a time with a fixed event vocabulary and feature engineering tied to a specific institution.
Our pipeline treats life events as one of three attribute categories within a unified profile, generated through the same LLM call and represented as free text over a much broader event vocabulary.
Different banks, event definitions, and cohorts (Section~\ref{sec:discussion}) make a direct numerical comparison with this line not apples-to-apples, and we do not stage one.

% ¶4 Evaluation methodology + schema roots.
%    LLM-judge for §5¶2 track (b) and §6.2¶4 only; calibration anchors ground earlier use of `prior` in §4¶5;
%    schema roots: kotler17 primary, smith56/wells71/cooper99/iab20 supporting; clustering tools one-line.
For the tag-based attribute evaluation of Section~\ref{sec:setup} and Section~\ref{sec:results}, the same LLM acts as a judge on the raw history and on the attribute-database input under the same prompt template, an instance of the LLM-as-a-judge methodology of \cite{zheng23_mtbench,liu23_geval}; the bias properties of this symmetric design are discussed in Sections~\ref{sec:setup} and~\ref{sec:discussion}.
The per-attribute prior $\prior{\freeattr}$ introduced earlier at Section~\ref{sec:method} rests on the verbalized-confidence formulation of \cite{tian23_just_ask} and the confidence-elicitation analysis of \cite{xiong24_confidence_elicitation}; Section~\ref{sec:results} tests its discriminative signal between positive and negative users rather than probability calibration.
The schema's three-category split (demographic, psychographic-behavioral, life-event) descends from the segmentation framework of \cite{kotler17_principles_marketing}, with roots in \cite{smith56_market_segmentation,wells71_aio,cooper99_inmates} and the industry-standard hierarchy of \cite{iab20_audience_taxonomy}; the clustering toolchain that turns free-text into a tag vocabulary follows standard practice (sentence-BERT~\cite{reimers19_sentence_bert}, UMAP~\cite{mcinnes18_umap}, HDBSCAN~\cite{campello13_hdbscan}) and is described in Section~\ref{sec:method}.

\section{Task Formulation}
\label{sec:formulation}

% ¶1 Input X_u and hybrid output Y_u = (Y^closed, Y^open) with prior \prior{d}
We are given each user's purchase or transaction history as a sequence of item names $\histX = (p_1, \dots, p_T)$, where each $p_i$ is the unprocessed text of one purchased item or transaction and time stamps are not consumed by the pipeline.
Our task is to produce, for each user, a hybrid profile $\profile = (\Yclosed, \Yopen)$ that combines a closed-set assignment with an open-set free-text description.
The closed-set side $\Yclosed$ records the user's assignment to a schema $\schema$ derived from the segmentation framework of~\cite{kotler17_principles_marketing} and organized into three attribute categories: seven demographic attributes $\schemadem$, four psychographic-behavioral attributes $\schemapb$, and eight life-event attributes $\schemale$, each with a mandatory \texttt{unknown} option for abstention.
The open-set side $\Yopen = \{\freeattr_1, \dots, \freeattr_M\}$ is a set of free-text attributes, short noun phrases or clauses such as ``education-focused parent of a school-age child,'' with $M \ge 0$.
Each free-text attribute $\freeattr_m$ carries a verbalized, unitless score $\prior{\freeattr_m} \in [0,1]$ from the LLM \cite{tian23_just_ask} — we call it the \emph{prior} of $\freeattr_m$ — indicating how strongly the attribute is expected to apply to users whose history matches the same inference unit (defined below).

\begin{figure}[t]
\centering
\includegraphics[width=0.75\columnwidth, trim={0 15pt 0 5pt}, clip]{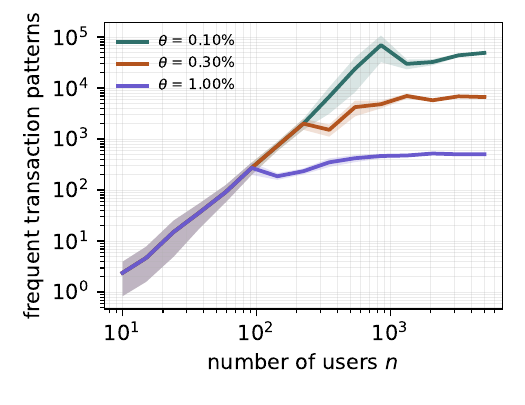}
\caption{Saturation of the unique-pattern count on the public Open e-commerce corpus~\cite{berke24_open_ecommerce} as the user count $n$ grows, plotted for three relative-frequency thresholds $\theta$ (share of \emph{users}, not transactions). For each $\theta$ the count stabilizes once $n$ is large enough for $\theta n$ to act as the binding threshold; at smaller $n$ a practical mining floor dominates instead. Item titles are kept raw rather than semantically abstracted; the shaded band shows the mean $\pm$ standard deviation across ten random user-orderings.}
\label{fig:saturation}
\end{figure}

% ¶2 Pattern-level decomposition + user aggregation g(.); sublinear cost argument; Fig.~\ref{fig:saturation}
%    #19 項目7 (R1-iv / R3-D3) の最終形（2026-09-07）: saturation の理由は推測形の 1 文のみ
%    （購買が人気商品に偏るため）。崩れる条件（stationarity caveat）の文は「レビューコメントに
%    固執しすぎ」として著者判断で削除——R3-D3 への明示応答は本文に無い。tail-cut コストの
%    §6 forward reference も削除済み（§6 は全圧縮の合計コストしか測らないため）。
Rather than running LLM inference per user, we infer attributes for transaction \emph{patterns} that recur across users.
A transaction pattern $\pattern$ is a set of one or more abstracted item names, mined from the corpus and deduplicated across users to obtain the unique-pattern set $\patternset$.
Because $\patternset$ is shared across users, the inference budget grows sublinearly in the user count.
Figure~\ref{fig:saturation} confirms this on the public Open e-commerce corpus~\cite{berke24_open_ecommerce}; the deployed system of Section~\ref{sec:deployment} operates well inside the saturated regime.
The saturation likely arises because user purchasing behavior is skewed toward popular items.
Finally, the per-user profile is formed by aggregating across the patterns matched by the user's history (Sections~\ref{sec:method} and~\ref{sec:setup}).

% ¶3 Evaluation targets vs schema; survey-item mapping; exclude Q-sexual-orientation
The schema $\schema$ above is the \emph{designed} attribute space, fixed by the segmentation framework rather than by the evaluation corpus.
A separate question is which of these attributes can be quantitatively measured against external ground truth.
We restrict the quantitative evaluation to attributes whose corresponding survey item exists in the Open e-commerce self-report (a subset of the demographic and life-event categories, mapping onto the survey's \texttt{Q-demos-*}, \texttt{Q-substance-use-*}, \texttt{Q-personal-*}, and \texttt{Q-life-changes} fields), and report the remainder qualitatively.
For each attribute in this aligned subset, the closed-set choice options are matched to the survey's response set so that the symmetric evaluation of Section~\ref{sec:results} maps both inputs into the same answer space; the full alignment is enumerated in Section~\ref{sec:setup}.
Sexual orientation is deliberately omitted from the schema: it is excluded from the scope of inference at design time rather than filtered after inference (Section~\ref{sec:discussion}).
A missing \texttt{Q-life-changes} response is treated as the negative ``nothing happened in 2021'' label so the life-event cohort is well-defined.

\section{Method}
\label{sec:method}

\subsection{Overview}
\label{sec:method:overview}

% ¶1 Pipeline overview (Fig.1); precision-vs-recall division of labor
The pipeline groups into three phases (Figure~\ref{fig:teaser}): \emph{Resolve} prepares item names, \emph{Profile} mines frequent transaction patterns and emits attributes per pattern with their priors, and \emph{Tag} clusters the free-text attributes into a queryable database.
Within the precision--recall trade-off, the pipeline is precision-first at the input stage (Resolve) and recall-first at attribute generation (Profile).
Resolve admits explicit abstention so items with no specific signal do not pollute later inference.
Profile surfaces plausible free-text candidates per category (between zero and several), favoring recall; the per-attribute prior lets downstream consumers restore precision at use time.

\subsection{Resolve: item-name preparation}
\label{sec:method:resolve}

% ¶2 Search decision, grounding, semantic abstraction
Resolve operates per item and consists of three components.
A per-item search-decision component lets the LLM, acting as a small agent over a single item, judge whether external grounding is required and abstain when the item carries no discriminative signal.
When grounding is requested, a web-search API returns text snippets, and a separate LLM call reads them into a structured record with a mandatory reference tag identifying the snippet it relied on.
The rewrite component then transforms the (possibly grounded) item name into an inference-ready form that drops brand, packaging, and other identifying noise while preserving content useful for attribute inference.
We call this rewrite \emph{semantic abstraction} rather than entity normalization because its goal is tractable downstream inference, not canonicalization of real-world references.

\subsection{Profile: per-transaction-pattern attributes}
\label{sec:method:profile}

Profile has three components: a non-LLM frequent-pattern mining step that produces the inference units, a joint LLM call that emits closed-set labels and free-text attributes for each pattern, and a per-attribute prior carrying the LLM's prevalence estimate.

% ¶3 Frequent pattern mining (non-LLM); patterns include single items and
%    multi-item combinations; relative-frequency exclusion; dedup -> unique-pattern set \patternset{}
\paragraph{Frequent pattern mining}
The mining step is the only non-LLM step in the pipeline.
A pattern is a set of one or more co-occurring item names within a user's time window: size-one patterns capture single discriminative items, and size-two-or-more patterns capture combinations that no constituent item alone supports.
A bundle of baby formula, diapers, and a crib in the same window, for example, supports a recent-childbirth inference far more strongly than any one of these items in isolation.
We mine patterns with the relative-frequency threshold $\theta$ of Section~\ref{sec:formulation}, deduplicate across users, and obtain the unique-pattern set $\patternset$.

% ¶4 Joint single-pass attribute inference (3 categories x 2 forms)
\paragraph{Joint attribute inference}
For each unique pattern $\pattern$, a single LLM call jointly produces, across all three attribute categories, the closed-set assignment $\Yclosed$ and the set of free-text attributes $\Yopen$ with their priors.
The two forms are emitted together by design.
The closed-set side maps each category to the small enumerated answer space of Section~\ref{sec:formulation} (each attribute with a mandatory \texttt{unknown} option), which anchors evaluation against external survey labels and stabilizes the downstream vocabulary.
The open-set side captures attributes that no closed enum can list in advance, such as \emph{education-focused parent of a school-age child}, which are the long-tail descriptions motivated in Section~\ref{sec:intro}.
Generating both in one pass keeps them conditioned on the same pattern context, avoids the cost and drift of a second call, and lets the comparison in Section~\ref{sec:results} decompose their contributions cleanly.
The same prompt structure handles single-item and multi-item patterns uniformly, injecting the pattern through a single \texttt{\{query\}} slot; the survey alignment of the closed-set options follows Section~\ref{sec:formulation}.
The open-set side is decoded recall-first: each category may emit zero to three candidates, with a category-specific canonical phrase frame (a noun phrase for demographic, a subject-verb clause beginning with \emph{user} for psychographic-behavioral, and a clause beginning with \emph{user or household recently} for life-event) that keeps downstream clustering robust to phrasing drift.

% ¶5 prior \prior{d}: verbalized prevalence; role in downstream precision control
\paragraph{Prior $\prior{\freeattr}$, the expected attribute match strength}
Each free-text attribute $\freeattr$ produced in Profile carries the prior $\prior{\freeattr}$ defined in Section~\ref{sec:formulation}, treated as independent across attributes for the same pattern.
The recall-first open-set side surfaces plausible-but-uncertain attributes; the prior is the single knob with which downstream consumers trade recall for precision at use time without re-prompting the LLM.
Concretely, \emph{education-investing parent} attached to a children's-tutoring purchase carries a high prior, while the same attribute attached to a generic stationery purchase carries a low one; thresholding on $\prior{\freeattr}$ drops weakly-supported attributes for a precision-sensitive task such as advertising targeting, while a lower threshold suits a recall-sensitive task such as audience discovery.
The same value weights the pattern-to-user aggregation when a user matches multiple patterns carrying the same attribute.
We make no claim that $\prior{\freeattr}$ is a calibrated probability; whether it functions as a discriminative signal between positives and negatives is a separate, post-hoc question examined in Section~\ref{sec:results}.
% Scope framing added 2026-09-01 (R4-4 / #19 項目8): calibration は out of scope として整理。
% 「できなかった」系の理由（prevalence 推定不可・粒度不一致）は書かない（morimura 方針 2026-08 in #19）。
Calibration is outside the scope of this study because our evaluation of the priors relies on their ordering rather than their accuracy as probabilities.
Applications requiring probabilities could calibrate the priors separately.

\subsection{Tag: attribute database}
\label{sec:method:tag}

% ¶6 Structuring free-text -> attribute database (SBERT -> UMAP -> HDBSCAN/KMeans -> tags)
Tag turns the per-transaction-pattern free-text attributes into a small queryable vocabulary and assembles the attribute database.
Within each of the three attribute categories, we embed the free-text attributes with a sentence encoder~\cite{reimers19_sentence_bert}, reduce dimensionality with UMAP~\cite{mcinnes18_umap}, cluster with HDBSCAN~\cite{campello13_hdbscan}, and ask the LLM to summarize each cluster into a short \emph{tag}~\cite{pham24_topicgpt}.
Tags are post-hoc relabelled or merged when the same concept fragments across nearby clusters.
Combining the tags with the closed-set labels from Profile yields a three-category \emph{attribute database} that records, for every user, the matched closed-set values, matched tags, and per-tag priors.
The database is indexed for natural-language queries (Section~\ref{sec:deployment}); its structuring quality is examined in Section~\ref{sec:results}.

\subsection{Implementation: prompt design}
\label{sec:method:implementation}

% ¶7 Prompt design — single-pass joint hybrid inference
Each LLM call is realized by a templated prompt that embeds the task definition, the schema where applicable, and the pattern under analysis as a single \texttt{\{query\}} slot; the LLM is invoked once per call on an on-premise Qwen3-family model (specific variants in Section~\ref{sec:setup}) and produces structured (JSON-like) output that the next stage consumes.
The inference template in Profile is the only one that combines closed-set and open-set generation, so we describe its anatomy explicitly.
It contains six elements in order: (i) a task description targeting the items in the pattern; (ii) one block per attribute category listing the closed-set choice enum with category-specific rules (no cross-category restatements, mandatory \texttt{unknown}, attribute-type-specific value formats); (iii) a format specification for the open-set side bounding each free-text attribute to a short atomic clause and capping the per-category count at three; (iv) an explicit prior-elicitation rule for the per-attribute prevalence among pattern purchasers in $[0,1]$; (v) worked few-shot examples for single-item and multi-item patterns; and (vi) a fixed-order JSON output schema for positional downstream parsing.
Beyond the joint single-pass generation motivated above, two design choices in this template are load-bearing.
Carrying the schema inside the prompt anchors the closed-set choices without training a separate classifier and lets the schema be specialized at deployment time without retraining (Section~\ref{sec:deployment}).
Eliciting the prior inside the same output structure, rather than in a follow-up call, makes the precision-recall knob available at no extra inference cost.
The other LLM calls (search routing, grounding read-out, and semantic abstraction in Resolve; cluster-naming in Tag) use task-specific templated prompts of the same structured form; malformed outputs are retried once.

% !TEX root = main.tex
% sec_setup.tex — Section 5: Experimental Setup
% Source: outline.md §5 (4 paragraphs)
% Page target: 1.0 page
%
% Conventions:
%   - "oracle" not "baseline" for the raw-history input (notation.md §3.5).
%   - "input" not "arm" (notation.md §4).
%   - "attribute database" for the DB as a whole; "tag" for clustered free-text elements.
%   - Track (a) selectable = ASYMMETRIC (DB side has no LLM at eval time; OR / weighted-median / mode aggregation).
%   - Track (b) tag-based  = SYMMETRIC (same LLM-as-judge processes both inputs).
%   - "symmetric" / "same LLM" wording is for track (b) ONLY.
%   - "merchant-name resolution" is §7 ONLY; never mention in §5.
%   - "prior" defined at §4.2¶5; symbol \prior{d}; used consistently.
%   - Code/data release is NOT mentioned in paper body (policy 2026-06-04).

\section{Experimental Setup}
\label{sec:setup}

% ¶1 Dataset and cohort
We evaluate the pipeline on Open e-commerce~\cite{berke24_open_ecommerce}, a public MIT-licensed corpus that pairs Amazon purchase histories with a self-reported survey on demographics, lifestyle, health, and recent life events.
It contains roughly $5{,}027$ users and $1.85$M purchases recorded between 2018 and 2022 over $97{,}801$ unique products.
We apply a single product-level cohort filter requiring each product to be purchased by at least three distinct users; products below this floor are dropped, after which $4{,}990$ users enter the selectable-attribute evaluation.
The tag-based evaluation additionally restricts to users with at least one positive survey label among the evaluated attributes and to users for whom all three input variants (oracle, hybrid, tags-only; introduced below) can be computed, leaving $3{,}781$ users.
Ground-truth labels are derived from the survey as follows.
For the five Yes/No items (cigarettes, marijuana, alcohol, diabetes, wheelchair) we map ``yes''/``stopped'' to $1$ and ``no'' to $0$, excluding other responses.
For the five life-event items (had child, became pregnant, moved, lost job, divorce) drawn from the \texttt{Q-life-changes} field, a non-response excludes the user from all five labels jointly, and a present response with no matching item becomes an explicit negative.
This asymmetric treatment of missing responses produces the gap in label availability between the two groups ($N{\approx}3{,}773$ vs.\ $N{=}1{,}636$); the implied cohort selection bias is discussed in Section~\ref{sec:discussion}.

% ¶2 Evaluation protocol: two tracks (selectable asymmetric, tag-based symmetric)
We measure information preservation by predicting Amazon survey attributes from two inputs: the raw transaction history, which we call the \emph{oracle} input, and the proposed attribute database.
The oracle carries strictly more information than any derived compressed representation, so it serves as an information-saturated upper reference rather than a competing baseline.
Each survey attribute is evaluated under one of two settings matched to the database's output form.
\paragraph{Selectable-attribute evaluation}
For attributes whose survey response space is ordinal or categorical (age, gender, income, education, relocation), the oracle-side prediction is emitted by the LLM acting as a predictor on the raw history.
The DB-side prediction aggregates the database's pattern-level closed-set choices across the user's matched patterns by an attribute-type-dependent rule (OR for relocation, weighted median over bin centers for age and income, mode for education and gender), with no LLM call at evaluation time; the setting is intentionally asymmetric.
We report one metric per attribute: macro-F1 (gender; mean F1 across classes), F1 (relocation), MAE in years (age), and ordinal MAE in bins (income, education).
\paragraph{Tag-based attribute evaluation}
For attributes whose survey response is a binary self-report (cigarettes, marijuana, alcohol, diabetes, wheelchair, and the five life-event items defined above), both inputs flow through the \emph{same} LLM acting as a judge $\judge$ under the same prompt template, emitting a probability per attribute.
We report per-attribute AUC, summarized by macro-AUC (the mean across the ten attributes).
The symmetric handling cancels position and verbosity bias in the judge~\cite{zheng23_mtbench,liu23_geval}; residual self-preference bias is acknowledged in Section~\ref{sec:discussion}.

% ¶3 Comparison scope and reference points
Rather than running a separate battery of external methods, we compare three inputs ordered by presentation logic: the \emph{oracle}, which feeds the raw transaction history to the same LLM; the \emph{hybrid} database input, which provides the user's clustered tags together with the closed-set labels assigned at the pattern level (the main proposed input); and the \emph{tags-only} database input, which omits the closed-set side as an ablation.
External non-LLM classifiers such as bag-of-products with logistic regression, or self-supervised transaction embeddings such as CoLES~\cite{babaev22_coles}, require a train/test split that does not align with our positive-label-only cohort, so an apples-to-apples re-run is deferred to future work (Section~\ref{sec:discussion}).
The oracle--hybrid comparison tests information preservation (Section~\ref{sec:results}); the hybrid--tags-only comparison decomposes the closed-set side's contribution (Section~\ref{sec:results}).

% ¶4 Models and reproducibility
We use Qwen3.5-27B with greedy decoding ($T=0$, sampling disabled) for both database construction and evaluation.
The bank deployment uses Qwen3-30B-A3B-Instruct-2507 on-prem (Section~\ref{sec:deployment}).
Free-text attributes are embedded with \texttt{plamo-embedding-1b}, reduced to eight dimensions with UMAP~\cite{mcinnes18_umap} (\texttt{n\_neighbors=20}, \texttt{min\_dist=0.1}, cosine), and clustered with HDBSCAN~\cite{campello13_hdbscan} (\texttt{min\_cluster\_ratio=0.01}, \texttt{min\_samples=20}, EOM, seed 3407); cluster tags are then generated by the same LLM.
Hyper-parameter ranges and selected values: $\minsup \in \{2,3,4,5\}$ ($3$); top-$K$ presented titles $\in \{200, 500, 1000\}$ ($500$).
Multi-item patterns are mined with FP-Growth up to four titles.
Database construction took approximately $33$ GPU-hours on a single NVIDIA A100 80GB (bfloat16, tensor parallel size $1$).
The closed-set schema is the three-category, $7{+}4{+}8$-attribute space of Section~\ref{sec:formulation}.
For attributes whose response space aligns with the Amazon survey, we set the closed-set options to the survey's response set; attributes outside the aligned subset retain the Kotler-derived options and are reported qualitatively (Section~\ref{sec:formulation}).
For the tag-based AUC table in Section~\ref{sec:results} we report $95\%$ percentile confidence intervals via stratified bootstrap ($B=1{,}000$ resamples, positive and negative pools resampled independently to preserve class counts); paired comparisons between inputs are summarized by wins across attributes and the Wilcoxon signed-rank $p$ value over the per-attribute point estimates.

\section{Results}
\label{sec:results}

\subsection{RQ1: Does the pipeline produce coherent, queryable attributes?}
\label{sec:results:rq1}

% ¶1 RQ1: qualitative examples (3 categories) — Amazon-side only
The pipeline produces coherent attributes across all three categories, including descriptions that no fixed schema encodes.
A children's picture book in the corpus yields the free-text attribute ``household with one or more young children under age 6'' at $\prior{d}=0.85$ together with the closed-set assignment \emph{having children: yes}; a baby-registry purchase yields ``recently celebrated a close friend or family member's childbirth'' at $\prior{d}=0.95$.
Such relational examples, where the gift buyer is distinct from the person expecting the child, lie outside any predefined consumer taxonomy~\cite{iab20_audience_taxonomy,kotler17_principles_marketing}.
In other cases the closed-set side returns \emph{unknown} on every attribute and only the free-text side carries signal: a personal-development workbook is described as ``recently initiated a mental health or therapy journey'' ($\prior{d}=0.45$); a memoir is described as ``undergoing a significant personal identity re-evaluation'' ($\prior{d}=0.55$).

% ¶2 RQ1: structuring of the attribute database (V4)
The Tag phase (Section~\ref{sec:method}) turns the per-transaction-pattern free-text outputs into a tag vocabulary.
Across the $97{,}801$ products in the corpus, the resulting vocabulary contains $76$ distinct tags after excluding one noise cluster per category: $23$ demographic, $23$ psychographic-behavioral, and $30$ life-event.
The mean number of tags per product is $1.05$, and $62{,}447$ products ($63.9\%$) carry at least one tag.
Tag coverage ranges from broad lifestyle segments (thousands of products each) down to rare life events (under $100$), within a single database: the largest tag, ``health-conscious home cooks and bakers prioritizing organic whole foods,'' covers $10{,}488$ products; ``engaged parents prioritizing early childhood educational and developmental play'' covers $3{,}845$; and the long-tail tag ``recently initiated new medication or medical treatment regimen'' covers $84$.
This breadth at per-product granularity is what makes the database queryable for downstream selection~\cite{wan24_tnt_llm}.

\subsection{RQ2 (central): Does the attribute database carry the history's signal?}
\label{sec:results:rq2}

% #19 項目6 は取り下げ（著者判断 2026-09-07）: 冒頭 lead-in は一度追加したが「文内に既に
% 書いてあるので不要」として削除。Wilcoxon / bootstrap CI の記載は ¶4 と Table 1 に既存。
% ¶3 §6.2 RQ2: Selectable-attribute evaluation
On selectable attributes the DB-side prediction is the deterministic aggregation of Section~\ref{sec:setup}; the oracle-side prediction is emitted by the LLM-as-a-predictor\footnote{Distinct from the \emph{LLM-as-a-judge} role of the tag-based evaluation below: the predictor emits a label or value directly, whereas the judge emits a probability of attribute membership.} on the raw history.
The DB input tracks the oracle to within $0.02$ on gender macro-F1 ($0.814 \to 0.794$) and income ordinal MAE ($1.252 \to 1.272$ bins), and increases age MAE by $0.58$ years ($8.22 \to 8.80$).
On relocation it substantially improves F1, from $0.128$ to $0.287$; the change is driven by a recall jump from $0.079$ to $0.318$ at a modest precision cost ($0.337 \to 0.262$), reflecting that the per-transaction-pattern aggregation flags more candidate movers than the LLM-as-a-predictor does on the raw history.
The DB ordinal MAE on education is also lower than the oracle ($0.815$ vs.\ $0.900$), but this reflects a center-clustering artifact of mode aggregation rather than improved ordinal discrimination: Spearman $\rho$ drops from $0.154$ to $0.038$, so the ordering signal is in fact lost.
On age and income the small change in ordinal MAE is consistent with compression at bin boundaries rather than collapse of the signal; on gender the macro-F1 drop of $0.020$ is similarly modest.

% ¶4 §6.2 RQ2: Tag-based attribute evaluation
On tag-based attributes, the same LLM-as-a-judge $\judge$ emits a probability per attribute for both the oracle and DB inputs; per-attribute AUC values with $95\%$ percentile bootstrap CI are reported in Table~\ref{tab:auc}.
At the aggregate level the hybrid DB input tracks the oracle within statistical noise: macro-AUC is $0.611$ for both the oracle and the hybrid input, and a Wilcoxon signed-rank test over the ten attribute-level point estimates yields $p=0.922$, so the two are indistinguishable.
On attributes that turn on specific products (had\_child $0.791 \to 0.743$, became\_pregnant $0.816 \to 0.780$, marijuana $0.598 \to 0.578$), the oracle is stronger, consistent with information lost when product names are compressed into tags.
On most other attributes the hybrid input's point estimate equals or exceeds the oracle's.
On \texttt{lost\_job} and \texttt{divorce} the per-attribute CIs cross the chance level $0.5$ under all three inputs, and no method achieves reliable discrimination on these two attributes.
These results support the view that the attribute database preserves the history's signal at the aggregate level.

\begin{table*}[t]
\centering
\caption{Tag-based attribute evaluation: per-attribute AUC with $95\%$ bootstrap CI, evaluated on $N=3{,}781$ users. $n_+$ = positive label count per attribute.}
\label{tab:auc}
\begin{tabular}{lrlll}
\toprule
attribute & $n_{+}$ & oracle & tags-only & hybrid (closed-set + tags) \\
\midrule
cigarettes      & $901$   & $0.523\ [0.502, 0.545]$ & $0.531\ [0.511, 0.552]$ & $0.531\ [0.511, 0.551]$ \\
marijuana       & $1{,}190$ & $0.598\ [0.579, 0.617]$ & $0.575\ [0.557, 0.593]$ & $0.578\ [0.558, 0.597]$ \\
alcohol         & $2{,}351$ & $0.533\ [0.513, 0.552]$ & $0.543\ [0.525, 0.561]$ & $0.551\ [0.533, 0.569]$ \\
diabetes        & $613$   & $0.603\ [0.577, 0.629]$ & $0.605\ [0.578, 0.631]$ & $0.615\ [0.590, 0.641]$ \\
wheelchair      & $98$    & $0.715\ [0.654, 0.771]$ & $0.675\ [0.613, 0.733]$ & $0.705\ [0.647, 0.763]$ \\
had\_child      & $159$   & $0.791\ [0.749, 0.829]$ & $0.733\ [0.690, 0.773]$ & $0.743\ [0.701, 0.781]$ \\
became\_pregnant& $145$   & $0.816\ [0.776, 0.851]$ & $0.761\ [0.713, 0.805]$ & $0.780\ [0.735, 0.823]$ \\
moved           & $1{,}085$ & $0.535\ [0.507, 0.564]$ & $0.493\ [0.463, 0.521]$ & $0.524\ [0.496, 0.555]$ \\
lost\_job       & $594$   & $0.501\ [0.473, 0.528]$ & $0.482\ [0.457, 0.508]$ & $0.517\ [0.488, 0.545]$ \\
divorce         & $63$    & $0.492\ [0.419, 0.573]$ & $0.536\ [0.458, 0.611]$ & $0.565\ [0.484, 0.647]$ \\
\midrule
macro (mean)    &         & $0.611$                 & $0.593$                 & $0.611$ \\
\bottomrule
\end{tabular}
\end{table*}

% ¶5 §6.2 RQ2: Hybrid effect (closed + tags vs tags only)
In the same harness, the hybrid input differs from the tags-only input only by including the pattern-level closed-set labels alongside the clustered tags (Section~\ref{sec:setup}).
The hybrid input outperforms tags-only on $9$ of $10$ attributes (the single loss is cigarettes with $\Delta=-0.000$, a tie at the reported precision), the macro-AUC gap is $\Delta=+0.018$, and the Wilcoxon signed-rank test gives $p=0.004$.
This is the only statistically significant input-level comparison in the harness, and it is consistent in direction across attributes.
The tags-only input itself does not differ significantly from the oracle (macro-AUC $0.593$ vs.\ $0.611$, $4$ wins out of $10$, $p=0.193$).
The closed-set side therefore contributes information that is not redundant with the tags rather than recovering from a degraded baseline.

\subsection{RQ3: Does \texorpdfstring{$\prior{d}$}{the prior} carry signal?}
\label{sec:results:rq3}

% ¶6 RQ3: Prior signal validity
We test whether the verbalized $\prior{d}$ separates users with positive vs.\ negative ground-truth labels on attributes where a database tag maps directly to a survey item.
Three attributes satisfy this condition: became\_pregnant, wheelchair, and diabetes.
For each we test whether the positive pool stochastically dominates the negative pool in $\prior{d}$ with a one-sided Mann--Whitney $U$ test, reporting Cliff's $\delta$ as the effect size.
Wheelchair shows the strongest separation, $p=2.3 \times 10^{-4}$ with $\delta=0.334$ (medium effect, $n_{+}=39$, $n_{-}=489$); diabetes is significant with a small-to-medium effect, $p=8.6 \times 10^{-3}$ and $\delta=0.221$ ($n_{+}=55$, $n_{-}=113$); on became\_pregnant the test is not significant, $p=0.17$ with $\delta=0.10$ ($n_{+}=34$, $n_{-}=205$), because $\prior{d}$ saturates near the upper end of its range for that attribute.
The verbalized $\prior{d}$ therefore carries discriminative signal in an attribute-dependent way.

\section{Real-World Deployment}
\label{sec:deployment}

% ¶1 Deployment context and goal
The pipeline is deployed at a major Japanese bank for marketing analysis, replacing a manual, expert-driven process.
Staff used to read transactions and write per-user descriptions by hand; this did not scale beyond a handful of curated audiences, and life-event inferences (e.g., users who recently paid a wedding venue or maternity clinic) were not attempted at all.
The system shifts marketing from predefined audience segments~\cite{liu19_ralm} to person-level retrieval: operators query the attribute database for users matching arbitrary natural-language descriptions, instead of designing segment boundaries in advance.
The pipeline runs on-premises on bank data that cannot be publicly released due to confidentiality restrictions.
We report only the operational design and reportable aggregate ratios.
The pipeline currently profiles on the order of tens of millions of users.

% ¶2 Deployed instance — pattern unit, schema specialization, merchant-name resolution
The deployment specializes the general formulation of Section~\ref{sec:method}.
Transaction patterns are pairs of (counterparty account name, transaction direction), where direction is outgoing or incoming payment.
Direction is part of the pattern because the same counterparty carries different meaning in each direction: payments to a music school identify a parent or learner, while income from the school identifies an instructor.
The closed-set schema is also specialized: where the public-data formulation of Section~\ref{sec:method} uses a three-category taxonomy (demographic, psychographic-behavioral, life-event), the deployed closed-set is seven occupation- and role-centric attributes (occupation, full-time work status, part-time work status, student status, retiree status, having children, education focus), paired with open-set free-text attributes and their per-attribute~$\prior{d}$.
A Japan-specific preprocessing step applies: \emph{merchant-name resolution} maps raw half-width katakana counterparty strings to canonical legal entity names\footnote{tantivy~\url{https://github.com/quickwit-oss/tantivy} (Rust full-text search) and gBizINFO~\url{https://info.gbiz.go.jp/} (Japan METI corporate database).}; this is unnecessary on the Amazon corpus, where product titles are already descriptive English text.
Patterns whose counterparty cannot be uniquely resolved to a single business (typically because the resolved name matches several companies under similar trade names) are excluded from inference, a precision-first omission revisited quantitatively below.

% ¶3 System metrics and the scalability mechanism
Pattern-level inference decouples cost from user count.
The support threshold is chosen so that retained patterns account for approximately $95\%$ of the original transaction log, compressing tens of millions of users to roughly $50{,}000$ high-frequency cross-user patterns.
This is a $\approx 600\times$ reduction in inference targets, and roughly $1/300$ the cost of a naive per-user pipeline that submits every user's history through the same LLM API once\footnote{Token counts and API price per token held constant.}.
After the precision-first filter introduced above removes unresolvable counterparties, the retained patterns cover $82.6\%$ of transactions in a representative production window; the gap from $95\%$ reflects this filter rather than a shortfall in mining.
With users averaging on the order of five transactions per six-month window, more than $99.9\%$ of users have at least one transaction matched by a tagged pattern\footnote{$1 - (1 - 0.826)^{5} \approx 0.9998$, treating transactions as independent; robust to the exact average since even $n=4$ yields $> 99.9\%$.}.
The same compression mechanism is characterized on public data in Section~\ref{sec:formulation} (Fig.~\ref{fig:saturation}): patterns plateau once user count exceeds $1/\theta$ for relative-frequency threshold $\theta$, and with $n$ in the tens of millions the deployment operates several orders of magnitude beyond that knee.
These metrics evidence deployability and efficiency; predictive accuracy is validated on public data (Section~\ref{sec:results}).

% ¶4 Emergent attribute database and direction asymmetry
The free-text side yields a vocabulary of about $149$ human-readable semantic tags (distinct from the $76$ Amazon tags in Section~\ref{sec:results:rq1}; different datasets).
The themes span occupation and professional roles, supply-chain participation, real-estate ownership, finance and insurance, medical care, and lifestyle and life-event descriptors.
Although the deployed closed-set is occupation- and role-centric, the open-set generates psychographic and life-event tags such as \emph{health-conscious self-care lifestyle}, \emph{homeowner adapting to a life-stage change}, and \emph{post-retirement asset and inheritance planning}, showing that the closed-set does not bound the granularity.
Transaction direction separates tags into two coherent roles: incoming yields producer/professional/supplier tags, outgoing yields consumer/investor/lifestyle tags (Table~\ref{tab:direction-split}).
In contrast to engineered life-event detectors over a small fixed event set~\cite{decaigny20_life_events}, life-event-like tags here emerge from the same single-pass inference, with no separate event-detection model.

\begin{table}[t]
\centering
\caption{Direction asymmetry in the deployed attribute database: dominant inferred role and example tags per transaction direction. Counts are tags whose strength in the indicated direction exceeds $80\%$, from a representative database snapshot; example tag names are shortened for presentation.}
\label{tab:direction-split}
\small
\begin{tabular}{p{0.20\linewidth}p{0.50\linewidth}r}
\toprule
Direction & Dominant role and example tags & Tags \\
\midrule
Incoming \par (user is paid) & \emph{Producer / Professional / Supplier} (e.g., IT engineer; supply-chain supplier; finance professional; specialized service provider; manufacturing or logistics worker) & $53$ \\
\addlinespace
Outgoing \par (user pays) & \emph{Consumer / Investor / Lifestyle} (e.g., ongoing insurance subscriber; financially-literate middle-aged retail investor; health-conscious self-care; education-investing parent) & $56$ \\
\bottomrule
\end{tabular}
\end{table}

% ¶5 End-to-end example
A single end-to-end example shows how the pipeline recovers meaning from an obfuscated merchant string and produces direction-dependent profiles.
The input is a half-width katakana counterparty string, omitted here to protect merchant identity.
After the web-grounding step in the Resolve phase, the simplified name reads as ``a company providing household services --- housekeeping, babysitting, and elderly-care support.''
The Profile phase then generates, for each direction, a ranked list of free-text attributes with priors; we report the top two for each direction without cherry-picking.
For outgoing transactions, the top profiles are \emph{busy dual-income parent using babysitting together with housekeeping} ($\prior{d}\approx 0.45$) and \emph{individual or family using elderly-care and housekeeping to reduce caregiving burden} ($\prior{d}\approx 0.25$).
For incoming transactions, the top profiles are \emph{field staff or professional employed by the household-services firm} ($\prior{d}\approx 0.50$) and \emph{per-job freelancer or sole proprietor performing similar work} ($\prior{d}\approx 0.25$).
The same counterparty thus yields distinct consumer- and worker-side profiles depending on direction.

\begin{figure*}[t]
\centering
\includegraphics[width=0.95\textwidth]{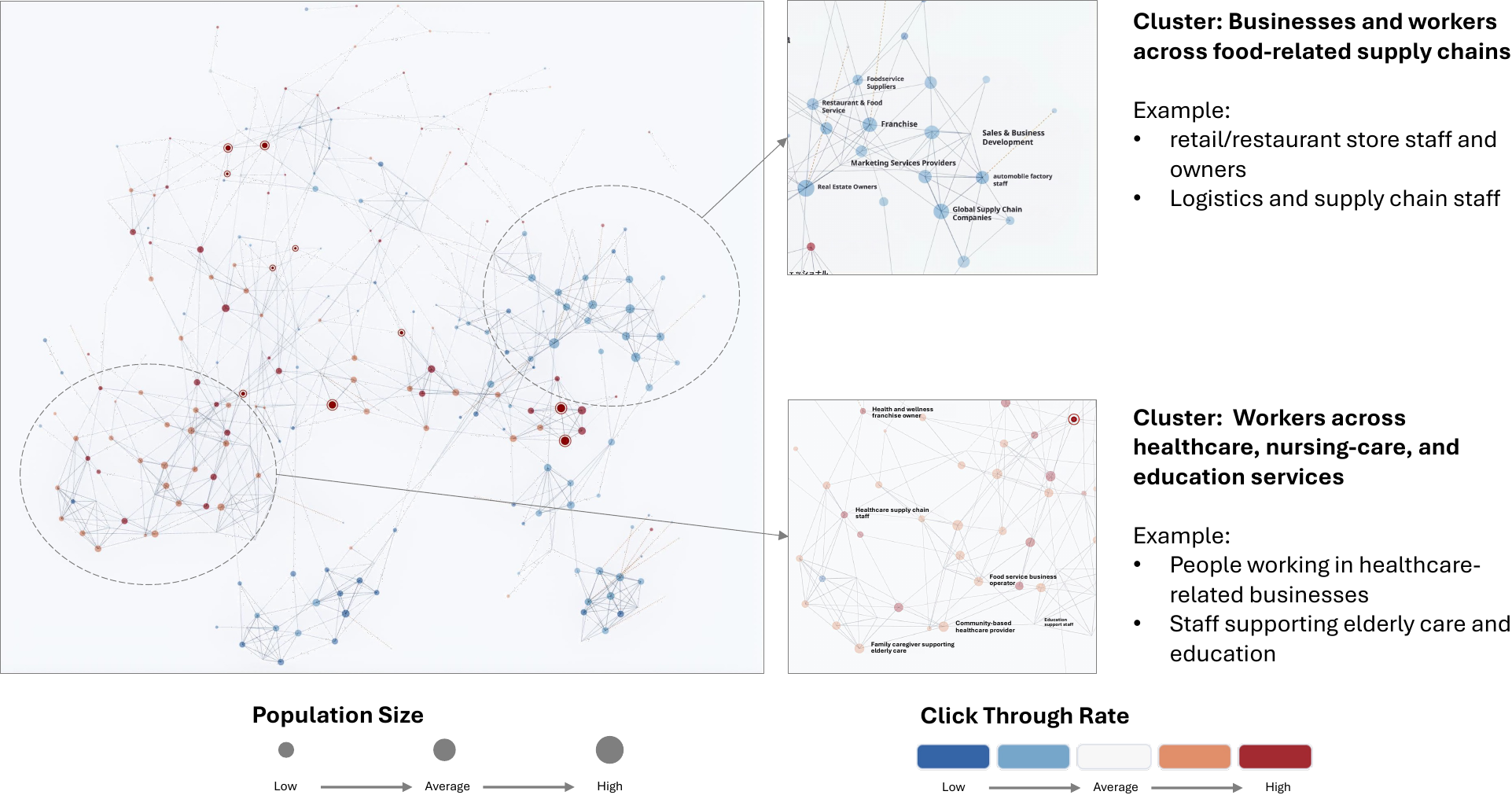}
\caption{Illustrative production view of the deployed attribute database used in routine marketing analysis. Nodes are semantic tags; node size reflects audience size (Population Size); node color reflects relative campaign response (Click Through Rate). The two zoom panels show example clusters: food-related supply chains, and healthcare, nursing-care, and education services. Values are noise-perturbed from real campaigns to protect commercial confidentiality; per-tag responses and cluster identities are not interpreted in the text.}
\label{fig:deployment-example}
\end{figure*}

% ¶6 Use case, figure, NL search, extensibility
In production the attribute database is queried as an index over users and tags rather than a fixed set of audience segments.
Figure~\ref{fig:deployment-example} shows a production view: each node is a semantic tag, node size reflects audience, and node color reflects relative campaign response.
Specific tag names, cluster identities, and campaign details are withheld to protect commercial confidentiality.
The two zoom panels show emergent clusters: businesses and workers across food-related supply chains, and healthcare, nursing-care, and education services; we do not interpret relative responses within them.
The database supports natural-language queries (Section~\ref{sec:method}), so operators retrieve users and tags by arbitrary descriptions without per-task training or labels.
The $\approx 149$ tags form an initial preset extended at operation time; the vocabulary has grown during the bank's use.
Sensitive-attribute handling and privacy implications are discussed in Section~\ref{sec:discussion}.

% !TEX root = main.tex
% sec_discussion.tex — Section 8: Discussion and Conclusion
% Source: outline.md §8 (4 paragraphs; former §9 merged as ¶4)
% Page target: 0.75 page
%
% Conventions:
%   - Dual-use / attribute-inference-attack discussion is mandatory (reviewers will probe).
%   - State Q-sexual-orientation exclusion and consent assumption explicitly.
%   - Limitations consolidate known reviewer risks: cohort bias, judge self-preference,
%     aggregation, life-event NL precedent, no external non-LLM baseline in this submission.
%   - No direct numerical comparison with decaigny20 / beyerdiaz24 (different banks/data) — say so.
%   - Future work is honest: aggregation ablations, non-LLM v-information comparison,
%     unimplemented tag-based audience targeting. NO insurance / financial-crime /
%     multi-company expansion lists (consistent with §7 no-expansion-list policy).
%   - No practitioner praise quotes.
%   - ¶4 conclusion (merged former §9): restate C1–C3 emphasizing formulation (§3) and
%     realization (§4) are BOTH methodological contribution. No "code & data public" sentence.
%   - "oracle" is §5/§6 terminology; if used here, treat as previously-defined.

\section{Discussion and Conclusion}
\label{sec:discussion}

% ¶1 Dual-use and privacy: explicit safeguards (Q-sexual-orientation exclusion,
%    consent assumption, on-prem), and the ad-delivery discrimination concern (ali19).
Inferring user attributes from purchase history shares structure with an attribute-inference attack on the same records.
We therefore treat privacy as a scope decision made at design time rather than as a property emerging from the method.
Sexual orientation is deliberately omitted from the schema of Section~\ref{sec:formulation} on ethical grounds, and the \texttt{Q-sexual-orientation} item is dropped from the Open e-commerce survey rather than treated as an evaluation target.
At the bank deployment of Section~\ref{sec:deployment}, the pipeline runs on-premises behind the institution's existing data boundary, so individual transactions never leave the bank; consent for marketing personalization is governed by existing customer agreements rather than by the pipeline.
Even attribute-only signals can still produce discriminatory exposure of opportunities such as housing or employment when fed into downstream ad-delivery systems~\cite{ali19_discrimination}; that responsibility lies with the targeting policy rather than with the attribute database, but it is a constraint operators should consider when defining tag-based audiences.

% ¶2 Limitations: cohort selection bias, residual self-preference, aggregation, life-event NL precedent,
%    no external non-LLM baseline this submission. Reference §5¶3 and §6.2¶4.
The evaluation of Section~\ref{sec:results} carries five limitations.
The tag-based cohort is restricted to users with at least one positive survey label, so the negative class is built from explicit non-responses rather than the full population, and effect sizes should be read as cohort-conditional (Section~\ref{sec:setup}).
The same LLM acts as judge on both inputs under matched prompt structures, which cancels position and verbosity bias but not self-preference bias~\cite{zheng23_mtbench}; the magnitude of any residual self-preference is not characterized here.
The pattern-to-user aggregation rule is fixed at a deterministic per-attribute-type choice without ablation; on education this fixed choice keeps ordinal MAE stable while losing the ordering signal (Section~\ref{sec:results}), and whether a different rule preserves ordinal structure better is left to future work.
A direct numerical comparison against engineered life-event detectors~\cite{decaigny20_life_events,beyerdiaz24_improved_life_event} is not possible here because those systems use different banks, event definitions, and a fixed discrete vocabulary; we therefore frame life-event coverage qualitatively (Section~\ref{sec:related}).
For the same reason of cohort and training-split mismatch, this submission omits an external non-LLM baseline (bag-of-products with logistic regression, or a self-supervised transaction embedding; see Section~\ref{sec:setup}).

% ¶2b User-specific vs shared-pattern (R3-D5 / #19 項目9): 著者指定の 2 文に短縮（2026-09-07）。
%    暗黙の仮定（tail の商品名は重要でない）＋仮定が破れる場合の最シンプルな対処（per-user 併用）。
%    future direction 等の文言は削除指示により無し。
By focusing on frequent patterns, the method may miss useful information from rare items.
For use cases where this matters, per-user inference could be added for users whose histories contain such items.

% ¶3 Future work: aggregation ablations, non-LLM symmetric v-information comparison,
%    unimplemented tag-based audience targeting. Honest — no readiness/imminence claims.
Five extensions follow.
A systematic ablation over aggregation rules (threshold, per-attribute-type choice, prior-weighted alternatives) would isolate how much of the per-attribute trend in Section~\ref{sec:results} is method-driven versus aggregation-driven.
A component-wise ablation of Resolve, Profile, and Tag together with an evaluation across different LLMs would clarify each stage's contribution and how much of the result depends on the specific model.
A quantitative evaluation of free-text attribute quality — e.g., the coherence between a tag and the products assigned to it, reported only qualitatively in Section~\ref{sec:results} — is likewise left to future work.
A non-LLM symmetric evaluation, in which a held-out classifier predicts the same labels from the raw history and from the attribute-database representation under a matched train/test split, would substitute a $\mathcal{V}$-information-style comparison for the LLM-as-a-judge protocol of Section~\ref{sec:setup} and address the residual judge-confounding concern raised above.
Tag-based natural-language audience targeting in the spirit of \cite{ramos24_nl_user_profiles,gao24_marketing_copilot,li25_consumer_segmentation_llm} is a natural use of the attribute database but is not implemented in the current deployment; we mention it as a future direction only, and likewise do not project the pipeline to adjacent settings beyond what is already running.

% ¶4 Conclusion (merged former §9): restate C1–C3 emphasizing formulation + realization
%    are both methodological contribution. One forward-looking sentence, no over-claim.
We have argued that re-casting LLM-based user-attribute inference from a per-user to a per-transaction-pattern problem is the load-bearing move (Section~\ref{sec:formulation}), and that its joint hybrid realization (Section~\ref{sec:method}) is part of the same methodological contribution rather than an implementation detail.
On the public Open e-commerce corpus the resulting attribute database tracks a raw-history oracle on macro-AUC within statistical noise on the tag-based evaluation, with consistent modest improvement once the closed-set side is added on top of the clustered tags (Section~\ref{sec:results}); deployed at a major Japanese bank, the same pipeline replaces a manual per-user reading workflow and operates well inside the plateau region the formulation predicts (Section~\ref{sec:deployment}).
It is the change of inference unit, not the choice of LLM or schema, that decouples cost from user count, and we expect the per-transaction-pattern unit to remain useful as natural-language attribute representations become a routine substrate for applied user-facing systems.

% Former §9 Conclusion is merged into sec_discussion.tex (§8 ¶4) per outline.md.
% McGill v2.0 reproducibility checklist is answered on the submission form
% (Applied Track allows form-only response; no in-paper section required).

\bibliographystyle{IEEEtran}
\bibliography{refs}

\end{document}